%% file: main.tex
\documentclass[nohyperref]{article}

\usepackage{microtype}
\usepackage{graphicx}
\usepackage{subfigure}
\usepackage{booktabs} 

\usepackage{hyperref}
\usepackage{nccmath}  
\usepackage{bbm}
\usepackage[flushleft]{threeparttable} 
\usepackage{multirow}

\iftrue
\newcommand{\cheng}[1]{{\color{teal}\textbf{cheng}: \textit{#1}}}
\else
\newcommand{\cheng}[1]{}
\fi

\iftrue

\else

\fi

\usepackage[accepted]{icml2022}

\usepackage{amsmath}
\usepackage{amssymb}
\usepackage{mathtools}
\usepackage{amsthm}

\DeclareMathOperator*{\argmin}{arg\,min}

\definecolor{Note_color}{rgb}{0.0, 0.0, 1.0}

\usepackage[capitalize,noabbrev]{cleveref}

\usepackage[symbol]{footmisc}

\theoremstyle{plain}

\theoremstyle{definition}

\theoremstyle{remark}

\usepackage[textsize=tiny]{todonotes}

\icmltitlerunning{DepthShrinker: A New Compression Paradigm Towards Boosting Real-Hardware Efficiency of Compact Neural Networks} 

\begin{document}

\twocolumn[
\icmltitle{DepthShrinker: A New Compression Paradigm Towards Boosting Real-Hardware Efficiency of Compact Neural Networks}





\begin{icmlauthorlist}
\icmlauthor{Yonggan Fu\footnotemark[2]{}}{rice}
\icmlauthor{Haichuan Yang}{meta}
\icmlauthor{Jiayi Yuan}{rice}
\icmlauthor{Meng Li}{meta}
\icmlauthor{Cheng Wan}{rice}
\icmlauthor{Raghuraman Krishnamoorthi}{meta}
\icmlauthor{Vikas Chandra}{meta}
\icmlauthor{Yingyan Lin}{rice}
\end{icmlauthorlist}

\icmlaffiliation{rice}{Department of Electrical and Computer Engineering, Rice University}
\icmlaffiliation{meta}{Meta Inc}

\icmlcorrespondingauthor{Yingyan Lin}{yingyan.lin@rice.edu}

\icmlkeywords{Machine Learning, ICML}

\vskip 0.3in
]



\printAffiliationsAndNotice{}  

\input{Sections/0-Abstract}

\input{Sections/1-Introduction}
\input{Sections/2-Related-Work}
\input{Sections/3-Motivation}

\input{Sections/4-Method}
\input{Sections/5-Experiment}
\input{Sections/6-Conclusion}

\bibliography{ref}
\bibliographystyle{icml2022}

\input{Sections/7-Appendix}


\end{document}

%% file: Sections/0-Abstract.tex
\begin{abstract}

Efficient deep neural network (DNN) models equipped with compact operators (e.g., depthwise convolutions) have shown great potential in reducing DNNs' theoretical complexity (e.g., the total number of weights/operations) while maintaining a decent model accuracy. However, existing efficient DNNs are still limited in fulfilling their promise in boosting real-hardware efficiency, due to their commonly adopted compact operators' low hardware utilization. In this work, we open up a new compression paradigm for developing real-hardware efficient DNNs, leading to boosted hardware efficiency while maintaining model accuracy. Interestingly, we observe that while some DNN layers' activation functions help DNNs' training optimization and achievable accuracy, they can be properly removed after training without compromising the model accuracy. Inspired by this observation, we propose a framework dubbed DepthShrinker, which develops hardware-friendly compact networks via shrinking the basic building blocks of existing efficient DNNs that feature irregular computation patterns into dense ones with much improved hardware utilization and thus real-hardware efficiency. Excitingly, our DepthShrinker framework delivers hardware-friendly compact networks that outperform both state-of-the-art efficient DNNs and compression techniques, e.g., a 3.06\% higher accuracy and 1.53$\times$ throughput on Tesla V100 over SOTA channel-wise pruning method MetaPruning. Our codes are available at: \href{https://github.com/facebookresearch/DepthShrinker}{https://github.com/facebookresearch/DepthShrinker}.

\end{abstract}

%% file: Sections/1-Introduction.tex
\section{Introduction}

\footnotetext[2]{Work done in collaboration between Meta Reality Labs and Rice EIC lab during internship}

Recent breakthroughs in deep neural networks (DNNs) have fueled a growing demand for deploying DNNs in real-world devices. However, the prohibitive complexity of DNNs stands at odds with the often constrained on-device resources. As such, many techniques aiming to boost DNNs' hardware efficiency have been developed, 
including pruning~\cite{han2015deep}, quantization~\cite{zhu2016trained,zhou2016dorefa}, and efficient DNN models ~\cite{MobilenetV1,efficientnetlite} leveraging compact operators (e.g., depthwise convolutions). 
Yet, the resulting DNN models from the above techniques mostly require dedicated DNN accelerators to achieve the desired hardware efficiency. 

In parallel, \textit{there exists a dilemma between the trends of efficient DNN design and modern computing platform advances}: while modern computing platforms (e.g., GPUs and TPUs) have consistently advanced to favor a higher degree of parallel computing, existing efficient DNN models often adopt light-weight operations that suffer from low hardware utilization and thus inferior achievable hardware efficiency.
For instance, depthwise convolutions~\cite{MobilenetV1}, commonly adopted in compact DNNs such as MobileNetV2 and EfficientNet, feature much more irregular computation patterns as compared to standard convolution layers, making it difficult to make good use of on-device resources due to their reduced data reuse opportunities and limiting existing efficient DNNs to unleash their theoretical potential ~\cite{chen2019eyeriss}. 
Therefore, there has been an increasing interest in developing more hardware friendly DNNs with improved hardware utilization to better leverage the power of parallelism in modern computing platforms~\cite{chen2018shallowing, elkerdawy2020filter,zhou2021evolutionary}.

To tackle the aforementioned gap between (1) the low hardware utilization of existing efficient DNNs and (2) the continuously increasing degree of computing parallelism of modern computing platforms,
we ask an intriguing question: ``\textit{How do we design efficient DNNs that can simultaneously enjoy both the powerful expressiveness of state-of-the-art (SOTA) efficient DNN structures and boosted parallel-computing capability of modern computing platforms}?" Inspired by RepVGG~\cite{ding2021repvgg}, which merges parallel branches to build decent single-branch networks, one natural thought is to merge consecutive layers into one single layer with dense computation patterns and improved hardware utilization. Nevertheless, it is non-trivial to merge layers along the depth dimension due to the associated activation functions, which are desired for introducing more non-linearity to empower the model capacity.

Interestingly, we observe that while some DNN layers' activation functions help DNNs' training optimization and thus achievable accuracy, they can be properly removed after training without compromising the model accuracy.
An exciting outcome is that the remaining consecutive linear operations between which the activation functions are removed can be merged into one single linear operation. \textbf{Notably, if the two activation functions in an inverted residual block~\cite{mobilenetv2}, the basic building block of SOTA efficient DNNs~\cite{mobilenetv2, tan2019efficientnet, wu2019fbnet, howard2019searching}, are removed, its two pointwise convolution layers and one depthwise convolution layer as well as the associated residual connection can be merged to one dense convolution with (1) a kernel of the same size as the original depthwise convolution and (2) the same number of input/output channels as the original inverted residual block}. Excitingly, the resulting dense convolution enjoys a much improved hardware utilization as compared to that of both the pointwise convolutions of kernel size $1\times1$ and depthwise convolution in the original inverted residual block, enabling the derived DNN to win boosted hardware efficiency while maintaining the original accuracy.

Driven by the above exciting discovery, we propose a new compression paradigm towards real-hardware efficient compact networks and make the following contributions:

\begin{itemize}

    \item We conduct experiments to show that the commonly adopted building blocks in existing efficient DNNs are inferior in hardware efficiency as compared to dense operations with the same theoretical complexity.

    \item Motivated by the above, we propose DepthShrinker that advocates \textbf{merging consecutive layers}, between which the activation functions are learned to be unimportant for inference, \textbf{into one single dense layer}. DepthShrinker's derived DNNs can largely leverage the high degree of parallelism in modern computing platforms and thus boost hardware efficiency while maintaining the original models' accuracy.

    \item DepthShrinker opens up a new perspective towards powerful and hardware-efficient DNNs, and can be viewed as some sort of soft layer pruning, in contrast to layer-wise pruning, i.e.,  \textit{merging} vs. \textit{hard pruning}.
    Notably, DepthShrinker delivers DNNs that outperform both SOTA channel- and layer-wise pruning techniques, e.g., a 3.06\% higher accuracy and 1.53$\times$ throughput on Tesla V100 over SOTA channel-wise pruning method MetaPruning~\cite{liu2019metapruning}. 

    \item Extensive experiments and ablation studies validate that DepthShrinker can \underline{(1)} largely push forward the frontier of DNNs' achievable accuracy-efficiency trade-off, and \underline{(2)} serve as an augmentation technique for boosting tiny DNNs' accuracy.

\end{itemize}

\label{sec:intro}

%% file: Sections/2-Related-Work.tex
\vspace{-0.5em}
\section{Related Works}
\label{sec:related-work}

\textbf{Efficient DNNs.}
Various efficient DNNs have been developed.
Early efficient DNNs mainly rely on human experts' manual design, e.g.,
MobileNets~\cite{MobilenetV1, mobilenetv2} boost model efficiency and accuracy trade-offs via depthwise convolution, which has become a standard operator for efficient DNNs.
In parallel, hardware efficient operators have been proposed~\cite{Shift, chen2020addernet} as alternatives for convolution. 
Thanks to the great success of neural architecture search (NAS)~\cite{zoph2016neural, zoph2018learning}, automated efficient DNN design via reinforcement learning~\cite{tan2019mnasnet, howard2019searching, tan2019efficientnet} and differentiable search~\cite{liu2018darts, wu2019fbnet, cai2018proxylessnas} have been proposed. 
However, existing efficient DNNs are still limited in their hardware efficiency due to the low hardware utilization of their basic building blocks, e.g., depthwise convolutions~\cite{MobilenetV1}.

\textbf{DNN compression techniques.}
Existing DNN compression techniques reduce the model complexity by pruning~\cite{han2015learning, han2015deep, wen2016learning, he2018amc, liu2019metapruning,he2017channel,he2020learning,he2019filter,dong2017learning}, quantization~\cite{courbariaux2015binaryconnect, courbariaux2016binarized, rastegari2016xnor,fu2020fractrain,fu2021cpt}, low-rank decomposition~\cite{yin2021towards,sainath2013low,nakkiran2015compressing}, or dynamic inference~\cite{teerapittayanon2016branchynet, wang2018skipnet,shen2020fractional}, while striving to maintain a decent accuracy.
Nevertheless, it is well known that general computing platforms (e.g., GPUs and CPUs) cannot fully benefit from DNN compression via low-bit quantization, low-rank decomposition, or dynamic inference in terms of hardware efficiency, and are still limited in fulfilling the efficiency improvement from pruning.

\begin{table*}[!t]
    \vspace{-0.5em}
\centering
\caption{Measured throughput of both the MobileNetV2 (including EfficientNet-Lite0) and ResNet families, as well as their corresponding dense counterparts on three commercial devices. All the reported numbers are real-device Frame-Per-Second (FPS).}
\resizebox{0.95\linewidth}{!}
{    
\begin{tabular}{cccccccc}
\toprule
\multirow{2}{*}{\textbf{Model}} & \multirow{2}{*}{\textbf{GFLOPs}} & \multicolumn{2}{c}{\textbf{Tesla V100 GPU}} & \multicolumn{2}{c}{\textbf{RTX 2080Ti GPU}} & \multicolumn{2}{c}{\textbf{TX2 Edge GPU}} \\
 &  & \textbf{Original} & \textbf{Dense} & \textbf{Original} & \textbf{Dense} & \textbf{Original} & \textbf{Dense} \\ \midrule
MobileNetV2 & 0.33 & 3088 & \textbf{12090 ($\uparrow$3.91$\times$)} & 2364 & \textbf{9351 ($\uparrow$3.96$\times$)} & 115 & \textbf{397 ($\uparrow$3.45$\times$)} \\
MobileNetV2-1.4 & 0.63 & 2127 & \textbf{8846 ($\uparrow$4.16$\times$)} & 1617 & \textbf{6869 ($\uparrow$4.25$\times$)} & 73 & \textbf{267 ($\uparrow$3.66$\times$)} \\
Efficientnet-Lite0 & 0.41 & 2731 & \textbf{11174 ($\uparrow$4.09$\times$)} & 2185 & \textbf{9577 ($\uparrow$4.38$\times$)} & 98 & \textbf{360 ($\uparrow$3.67$\times$)} \\ \midrule \midrule
ResNet-50 & 4.14 & 1079 & \textbf{2182 ($\uparrow$2.02$\times$)} & 874 & \textbf{1862 ($\uparrow$2.13$\times$)} & 45 & \textbf{53 ($\uparrow$1.18$\times$)} \\
ResNet-101 & 7.88 & 642 & \textbf{1509 ($\uparrow$2.35$\times$)} & 538 & \textbf{1279 ($\uparrow$2.38$\times$)} & 28 & \textbf{44 ($\uparrow$1.57$\times$)} \\
ResNet-152 & 11.62 & 449 & \textbf{1082 ($\uparrow$2.41$\times$)} & 378 & \textbf{917 ($\uparrow$2.43$\times$)} & 19 & \textbf{30 ($\uparrow$1.58$\times$)} \\ \bottomrule
\end{tabular}
}
\label{tab:profiling}
\vspace{-1em}
\end{table*}

\textbf{Layer-wise pruning.}
The most relevant work to DepthShrinker is layer-wise pruning~\cite{chen2018shallowing, elkerdawy2020filter,zhou2021evolutionary, xu2020layer}, which prunes an entire layer/block motivated by the fact that pruning a layer is more effective in reducing hardware latency~\cite{xu2020layer} compared with channel-wise pruning. Specifically,~\cite{chen2018shallowing, elkerdawy2020filter} prune layers based on their proposed criteria; while~\cite{zhou2021evolutionary} and \cite{xu2020layer} determine which layers/blocks to be pruned via evolutionary search and differentiable optimization, respectively. However, aggressive layer pruning inevitably suffers from non-trivial accuracy drops under large compression ratios due to the difficulty in restoring the pruned models' accuracy. Instead of hard pruning, our DepthShrinker merges consecutive linear operations into one dense operation after training, and can win both accuracy and hardware efficiency thanks to (1) its better maintained model expressiveness and (2) the high utilization and thus lower latency of the merged dense operation.

%% file: Sections/3-Motivation.tex
\begin{figure}[t!]
\centering
\includegraphics[width=0.98\linewidth]{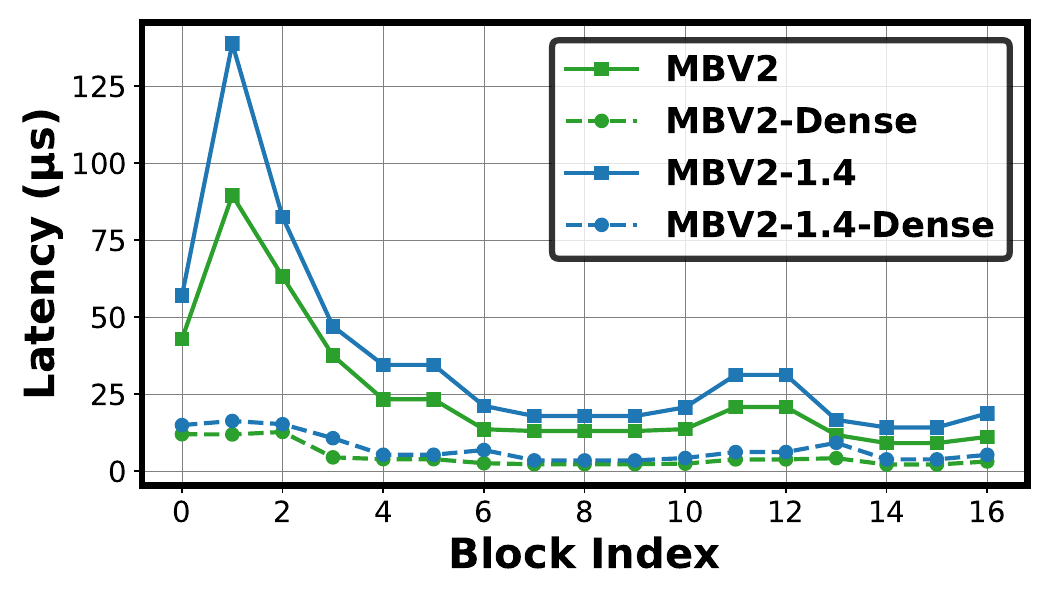}
\vspace{-1.5em}
\caption{Visualizing the block-wise latency of the inverted residual blocks (a total of 17) in MobileNetV2/MobileNetV2-1.4 (\textit{solid lines}) and their corresponding dense convolutions (\textit{dashed lines}) on an RTX 2080Ti GPU. ``MBV2" denotes MobileNetV2.}
\label{fig:profile_motiv}
\vspace{-1.5em}
\end{figure}

    \vspace{-0.5em}
\section{Motivating Inspiration and Observations}
\label{sec:motivation}

\subsection{Inspiration Drawn from Previous Works}
\label{sec:inspiration}
\textbf{Shallow networks with a higher utilization favor real-hardware efficiency.} Recent works~\cite{elkerdawy2020filter, xu2020layer} show that shallow networks favor a higher degree of parallel processing and thus higher hardware utilization, leading to better real-hardware efficiency on modern computing devices, e.g., GPUs, over their deeper counterparts with a comparable computational cost; this is also further validated by our profiling experiments in Sec.~\ref{sec:profiling}. Nevertheless, existing shallow networks are still not able to approach the accuracy of their deep counterparts, motivating us to innovate shallow networks with boosted accuracy.

\textbf{Linear operations in DNNs can be merged.} It is known that linear operations can be properly merged.
RepVGG~\cite{ding2021repvgg} shows that linear operations in parallel branches of a DNN can be merged to deliver competitive single-branch networks. Inspired by this, one natural thought for building powerful compact networks is to merge consecutive layers in a SOTA DNN to reduce the model depth. However, layers cannot be directly merged along the depth dimension due to the non-linear activation functions. 
This motivates us to question ``whether some activation functions can be properly removed for inference".

\textbf{The role of activation functions.} We hypothesize that the answer to the above question is positive based on existing DNN compression~\cite{deep_compression,jacob2018quantization} and training ~\cite{zhou2020go,cai2021network} works, which show that DNNs' higher complexity benefits training but can be trimmed down during inference without hurting the accuracy.
Specifically, iterative pruning~\cite{deep_compression} and quantization-aware training~\cite{jacob2018quantization} train DNNs with their original complexity and then sparsify/quantize the model for inference without hurting the accuracy; Meanwhile,~\cite{zhou2020go,cai2021network} augment DNNs via expanding their widths during training for improved accuracy, while the models during inference remain the same. One inspiration from these prior arts is that activation functions can be viewed as one specific model dimension in enhancing DNNs' complexity and expressiveness, and thus some might be properly removed after training without hurting the accuracy.

    \vspace{-0.5em}
\subsection{Motivating Observations from Real-device Profiling }
\label{sec:profiling}
\vspace{-0.5em}

Since our work is hardware-driven and aims to improve real-device efficiency instead of theoretical ones, we conduct extensive real-device profiling experiments to validate our hypothesis and to gain better understandings of the design space, and they are summarized in this subsection.

\textbf{Key hypothesis/motivation.} While the commonly used bottleneck blocks~\cite{he2016deep} and more efficient inverted residual blocks~\cite{mobilenetv2} have shown impressive theoretical efficiency and accuracy trade-offs, their real-device efficiency is inferior as compared to their dense counterparts under the same computational complexity, due to their more irregular computation patterns that cause reduced data reuse and lower hardware utilization.

\textbf{Experiment setup.} In our profiling, we replace \textit{each building block} in both the MobileNetV2~\cite{mobilenetv2} (including EfficientNet-Lite0~\cite{efficientnetlite}) and ResNet~\cite{he2016deep} families with \textit{one dense convolution layer} (1) of the same kernel size as the second convolution layer of each block, which is the only convolution with a kernel size larger than $1\times1$ within a bottleneck/inverted residual block and (2) with a scaled number of channels to \textbf{maintain the same  floating-point operations (FLOPs) as the original block}. 
We summarize the real-device throughput of both these two network families featuring two different types of basic building blocks and their dense counterparts on ImageNet with a resolution of 224 $\times$ 224 in Tab.~\ref{tab:profiling}.

\textbf{Considered devices and measurement settings.} We consider three commercial devices, including (1) NVIDIA Tesla V100 GPU~\cite{v100}, (2) NVIDIA RTX 2080Ti GPU~\cite{rtx2080ti}, and (3) Jetson TX2 Edge GPU~\cite{edgegpu}, to cover both Desktop  and edge GPUs. We adopt a batch size of 128 for the first two devices, following~\cite{ding2021repvgg}, and 64 for the last device, and Frame-Per-Second (FPS) as the efficiency metric.

\begin{figure*}[t!]
\centering
\includegraphics[width=0.89\linewidth]{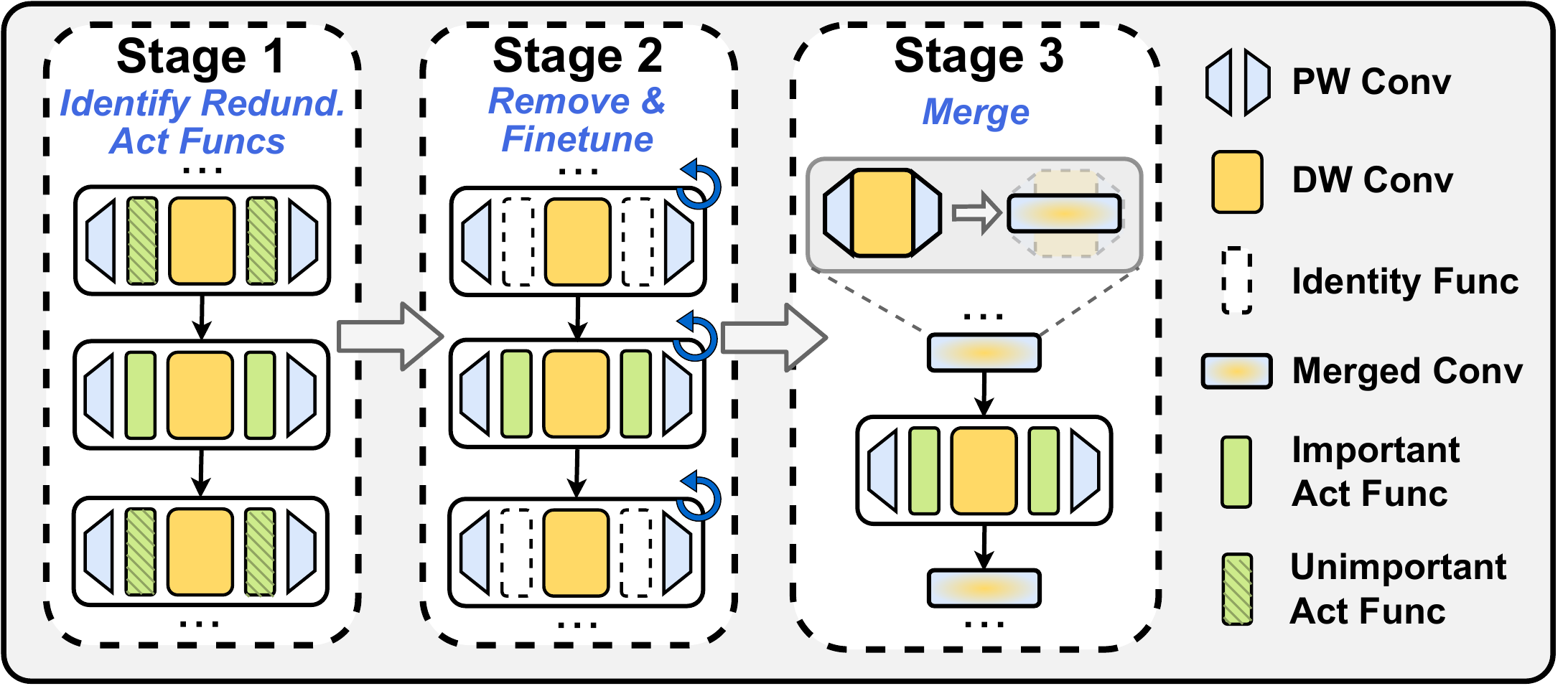}
\vspace{-0.5em}
\caption{Overview of our DepthShrinker framework and its three-stage design. ``PW" and ``DW" denote pointwise/depthwise convolutions, respectively. During merging, we merge the two pointwise convolutions and one depthwise convolution in blocks whose activation functions are removed, into one \textit{dense} convolution.
}
\label{fig:overview}
\vspace{-1.3em}
\end{figure*}

\textbf{Results and analysis.} As shown in Tab.~\ref{tab:profiling}, we can observe that \underline{(1)} the dense counterparts consistently achieve a higher throughput against the original networks with the same FLOPs, regardless of the model families and profiling devices. Specifically, the dense convolution counterparts on top of the MobileNetV2 family boost the throughput on a Tesla V100/RTX 2080Ti GPU and TX2 Edge GPU by 3.91$\times$ $\sim$4.38$\times$ and 3.45$\times$ $\sim$3.67$\times$, respectively; and similarly, they increase the throughput on a Tesla V100/RTX 2080Ti GPU and TX2 Edge GPU by 2.02$\times$ $\sim$2.43$\times$ and 1.18$\times$ $\sim$1.58$\times$, respectively, on top of the ResNet family.
To further understand this, we also visualize the block-wise latency of MobileNetV2/MobileNetV2-1.4 on a RTX 2080Ti GPU in Fig.~\ref{fig:profile_motiv}, including the latency of both the original block and the corresponding dense convolution. It shows that the dense counterpart for each block can consistently reduce the latency by up to 88.2\%. 

This set of profiling experiments indicates that \underline{(1)} replacing the commonly adopted building blocks with dense operations of the same FLOPs can notably boost real-device efficiency, thanks to the improved utilization of hardware resources; \underline{(2)} The throughput improvement is more notable on top of the MobileNetV2 family than the ResNet family, because depthwise convolutions in the former and widely adopted in existing efficient DNNs introduce more irregular computation patterns and thus we see a  more pronounced improvement after replacing them with dense ones; and \underline{(3)} throughput improvement is consistently observed across different devices, and is larger on the Tesla V100/2080Ti GPU than the TX2 Edge GPU since the former has a higher degree of parallel-processing that favors the achievable throughput of the dense counterparts, indicating an even larger efficiency improvement of our DepthShrinker, in line with more parallelism trend of modern AI-driven computing platforms.


\textbf{Remark.} The hardware utilization and thus real-device efficiency improvements 
are mainly attributed to two perspectives: \underline{(1) from the operation perspective}, irregular operations have less data reuse opportunities and thus require more data movement costs~\cite{chen2019eyeriss}, e.g., in our profile experiment, both the standard convolutions with a kernel size of $1\times1$ and the depthwise convolutions are replaced with dense convolutions with a larger kernel size of $3\times3$, leading to more data reuse opportunities and less data movements under the same FLOPs; and \underline{(2) from the depth/width trade-off perspective}, replacing a building block with one dense convolution of the same FLOPs would both shallow and widen the original network, and thus favor a higher utilization when running on modern computing platforms featuring an increasingly high degree of parallelism. We further study the independent impact of the above perspective \underline{(2)} in the Appendix.~\ref{app:profile}.

%% file: Sections/4-Method.tex
\section{The Proposed DepthShrinker Framework}
\label{sec:method}
\subsection{Overview}
\label{sec:method_overview}
\textbf{Key idea.} DepthShrinker aims to develop real-hardware efficient DNNs favoring high hardware utilization by removing redundant activation functions and then merging the resulting consecutive linear operations. The key idea is that by removing the two activation functions within an inverted residual block~\cite{mobilenetv2}, i.e., the basic building block in most efficient DNNs, the entire block can then be merged into one dense convolution layer with the same kernel size as the original block's depthwise convolution and the same number of  input/output channels as the original block. The exciting outcome is much improved real-hardware efficiency as profiled in Sec.~\ref{sec:profiling}.

\textbf{Framework overview.} To achieve the above aim, two nontrivial challenges exist:  which activation functions to be removed and how to restore the accuracy with fewer remaining activation functions after the removal. To tackle these, DepthShrinker built on top of SOTA efficient DNNs integrates a three-stage effort as shown in Fig.~\ref{fig:overview}:
\underline{(1)} identify redundant activation functions, \underline{(2)} remove the identified activation functions and fine-tune the resulting DNNs from stage (1), and \underline{(3)} merge consecutive layers between which the activation functions are removed to deliver the final networks. Note that we apply our DepthShrinker on top of publicly available pretrained models, following common practice in model compression~\cite{he2018amc,wen2016learning,han2015deep,jacob2018quantization}.

  \vspace{-0.5em}
\subsection{Stage 1: Identify Redundant Activation Functions}
\label{sec:method_search}
To identify unimportant activation functions, predefined criteria like those for layer-wise pruning~\cite{li2016pruning,wen2016learning} may not be appropriate as activation functions of different layers are coupled, e.g., removing preceding activation functions might change the feature distributions of the following layers. Therefore, we propose a differentiable search method to learn the importance of all activation functions considering their joint influence, as inspired by recent pruning works~\cite{kang2020operation,ramanujan2020s,fu2021drawing}.

\textbf{Search method overview.} 
Our search method assigns a learnable mask $m \in \mathbb{R}^{N}$ ($N$ is the total number of activation functions) to all activation functions, serving as a proxy of the activation functions' importance score. When updating $m$ during search, the coupling effect among different activation functions should be considered, while ensuring that the sparsity of $m$ is sufficiently high, e.g., higher than $(1-k/N)$ where $k$ is the number of remained activation functions, to satisfy the target efficiency after the merging stage (see sec.~\ref{sec:method_merge}). In DepthShrinker, the search method jointly learns the mask $m$ together with model weights $\theta$.

\textbf{Search method formulation.} Deriving the optimal $\theta$ and $m$ is essentially a bi-level optimization problem~\cite{liu2018darts}. In DepthShrinker, we approximate it as a one-level optimization formulation~\cite{xie2018snas,hu2020dsnas,bi2020gold} to jointly update $\theta$ and $m$:

\useshortskip
\begin{equation} \label{eq:search_objective}
    \argmin_{\theta, m} \,\, \sum_{i} \ell(\hat{y}_{\theta,m}(x_i), y_i) \,\,\,\,\, \text{s.t.} \,\,\, \|m\|_0  \leqslant k
\end{equation} 

where $\ell$ is the loss function, $x_i$ and $y_i$ are the $i$-th input and label pair, and
$\hat{y}_{\theta,m}(x_i)$
is the predicted label under the parameters $\theta$ and activation mask $m$. 
To achieve the target sparsity in $m$, we impose an $L_{0}$ constraint on $m$ via activating only its top $k$ elements during forward. 
Specifically, we adopt a binary mask $\hat{m}\in \{0,1\}^{N}$ to approximate the top $k$ elements of $m$ using 1 and 0 otherwise during forward, while all the elements in $m$ are updated via straight-through estimation~\cite{bengio2013estimating} during backward.
In particular, the forward function can be formulated as:

\useshortskip
\begin{equation}
f^{(l)}_{\theta,\hat{m}}(\cdot)\coloneqq(\hat{m}_l\cdot\sigma+(1-\hat{m}_l)\cdot\mathbbm{1})\circ\mathcal{T}_{\theta_l}\circ f^{(l-1)}_{\theta,\hat{m}}(\cdot)\label{eq:forward}
\end{equation}
\vspace{-1.5em}

where $f^{(l)}_{\theta,\hat{m}}$ is the network function for the first $l$ layers, $\circ$ is the operator of function composition, $\sigma$ and $\mathbbm{1}$ denote an activation function and identity mapping, respectively, and $\mathcal{T}_{\theta_l}$ is a transformation  (e.g., convolution or other linear operations) parameterized by $\theta_l$.
The binary mask $\hat{m}_l$ in Eq.~(\ref{eq:forward}) guarantees that an activation function in the $l$-th layer is either fully enabled or disabled. 
During backward, we directly pass the gradients of the binary mask $\hat{m}$ to $m$, i.e., $\frac{\partial \ell}{\partial m}\approx\frac{\partial \ell}{\partial \hat m}$.
Since only the activation functions corresponding to the top $k$ values of mask $m$ participate in the forward process, activation functions with larger $m$ values are more likely to be kept, thus larger $m$ values after training indicate higher importance.

\textbf{Search method implementation.} Our search method makes two settings: (1) block-wise shrink and (2) latency-aware decay on $m$. For the former, since we aim to merge the whole inverted residual blocks into one dense convolution, we share the mask values of $m$ corresponding to the two activation functions in one block, i.e., both of the two activation functions are either removed or kept. For the latter, we additionally add an $L_1$ decay on each element of $m$ weighted by the corresponding block's latency during search to penalize the importance of the costly blocks. 
Note that in this work we directly adopt the pre-measured latency on RTX 2080Ti GPU during search, and recognize that it is straightforward to make it platform-aware for further boosting the efficiency at the cost of a longer search time.

\subsection{Stage 2: How to Fine-tune}
\label{sec:method_ft}
After the above search process, the least important activation functions with the smallest $m$ are removed, and fine-tuning is performed to restore the accuracy. The following solutions have been proposed and validated: 

\textbf{Adding additional activation functions for free.} 
There is no nonlinear function following the dense convolutions after the merge stage, since an inverted residual block~\cite{mobilenetv2} contains only two activation functions. 
To boost the achieved accuracy, we additionally add an activation function (i.e., ReLU6 in this work) after each merge  convolution, which incurs a negligible hardware cost.

\textbf{Self-distillation.}
In DepthShrinker, we can optionally enable a self-distillation mechanism during fine-tuning, i.e., conducting knowledge distillation~\cite{hinton2015distilling} under the guidance of the original network with all activation functions on to further boost the derived network's accuracy. Note that we only assort to the original network as the teacher without introducing extra models.

\begin{table*}[!t]
  \vspace{-0.5em}
\centering
\caption{Benchmark DepthShrinker with SOTA channel-wise pruning method MetaPruning~\cite{liu2019metapruning} and uniform pruning on MobileNetV2-1.4@ImageNet in terms of FPS measured on three devices. All baseline accuracies are their reported ones~\cite{liu2019metapruning}.}
\resizebox{0.9\linewidth}{!}
{    
\begin{tabular}{cccccc}
\toprule
\textbf{Model} & \textbf{Acc (\%)} & \textbf{MFLOPs} & \textbf{Tesla V100} & \textbf{RTX 2080Ti} & \textbf{TX2} \\ \midrule
MBV2-1.4 & 75.30 & 630 & 2127 ($\uparrow$1.00$\times$) & 1617 ($\uparrow$1.00$\times$) & 73 ($\uparrow$1.00$\times$) \\ \midrule
MetaPruning-1.0$\times$ & 73.20 & 332 & 3159 ($\uparrow$1.49$\times$) & 2527 ($\uparrow$1.56$\times$) & 115 ($\uparrow$1.58$\times$) \\
\textbf{MBV2-1.4-DS-A} & 74.65/75.29/75.65 & 519 & 3827 ($\uparrow$1.80$\times$) & 2881 ($\uparrow$1.78$\times$) & 134 ($\uparrow$1.84$\times$) \\
\textbf{MBV2-1.4-DS-B} & 73.67/74.8/75.13 & 502 & 4778 ($\uparrow$2.25$\times$) & 3356 ($\uparrow$2.08$\times$) & 163 ($\uparrow$2.23$\times$) \\
\textbf{MBV2-1.4-DS-C} & 73.38/74.55/74.91 & 492 & 4597 ($\uparrow$2.16$\times$) & 3537 ($\uparrow$2.19$\times$) & 159 ($\uparrow$2.18$\times$) \\ \midrule \midrule
Uniform-0.65$\times$ & 67.20 & 182 & 4004 ($\uparrow$1.88$\times$) & 3147 ($\uparrow$1.95$\times$) & 161 ($\uparrow$2.21$\times$) \\
MetaPruning-0.65$\times$ & 71.70 & 160 & 4336 ($\uparrow$2.04$\times$) & 3691 ($\uparrow$2.28$\times$) & 179 ($\uparrow$2.45$\times$) \\
\textbf{MBV2-1.4-DS-D} & 72.51/73.93/74.50 & 484 & 5560 ($\uparrow$2.61$\times$) & 3926 ($\uparrow$2.43$\times$) & 184 ($\uparrow$2.52$\times$) \\
\textbf{MBV2-1.4-DS-E} & 72.20/73.85/74.43 & 474 & 5317 ($\uparrow$2.50$\times$) & 4175 ($\uparrow$2.58$\times$) & 179 ($\uparrow$2.45$\times$) \\ \midrule \midrule
Uniform-0.35$\times$ & 54.60 & 68 & 6266 ($\uparrow$2.95$\times$) & 5607 ($\uparrow$3.47$\times$) & 263 ($\uparrow$3.60$\times$) \\
MetaPruning-0.35$\times$ & 64.50 & 52 & 7044 ($\uparrow$3.31$\times$) & 6938 ($\uparrow$4.29$\times$) & 377 ($\uparrow$5.16$\times$) \\
\textbf{MBV2-1.4-DS-F} & 67.56/69.04/70.13 & 415 & 10804 ($\uparrow$5.08$\times$) & 7687 ($\uparrow$4.75$\times$) & 344 ($\uparrow$4.71$\times$) \\ \bottomrule
\end{tabular}
}
\label{tab:exp_metapruning}
\vspace{-1em}
\end{table*}

\subsection{Stage 3: How to Merge}
\label{sec:method_merge}
After fine-tuning the resulting network with unimportant activation functions removed, the final step is to merge adjacent linear operations (e.g., convolutional/fully-connected, average pooling, or batch normalization layers).

\textbf{Merging two adjacent layers.}  Without loss of generality, here we consider two adjacent convolution layers with an input feature $X\in\mathbb{R}^{H_1\times W_1\times c_1}$, intermediate feature $Z\in\mathbb{R}^{H_2\times W_2\times c_2}$, output feature $Y\in\mathbb{R}^{H_3\times W_3\times c_3}$, and kernel weights $K^{(1)}\in\mathbb{R}^{d_1\times d_1\times c_1\times c_2}$ and $K^{(2)}\in\mathbb{R}^{d_2\times d_2\times c_2\times c_3}$ for the first and second layers, between which the activation is removed, and assume:
\begingroup
\setlength{\abovedisplayskip}{3pt}
\setlength{\belowdisplayskip}{3pt}
\begin{align}
Z_{m,n,s}&=\sum_{i=0}^{d_1-1}\sum_{j=0}^{d_1-1}\sum_{r=0}^{c_1-1}X_{m-i,n-j,r}K^{(1)}_{i,j,r,s} \\
Y_{m,n,t}&=\sum_{i=0}^{d_2-1}\sum_{j=0}^{d_2-1}\sum_{s=0}^{c_2-1}Z_{m-i,n-j,s}K^{(2)}_{i,j,s,t} 
\end{align}
\endgroup
The above two layers can be merged into one single layer. \\Assuming the stride of both layers $s_1$ and $s_2$ is 1, we have: 
\begingroup
\setlength{\abovedisplayskip}{3pt}
\setlength{\belowdisplayskip}{3pt}
\begin{align}
Y_{m,n,t}=\sum_{i=0}^{d-1}\sum_{j=0}^{d-1}\sum_{r=0}^{c_1-1}X_{m-i,n-j,r}K_{i,j,r,t}
\end{align}
\endgroup
\noindent where $d=d_1+d_2-1$ and

\begingroup
\setlength{\abovedisplayskip}{0pt}
\setlength{\belowdisplayskip}{0pt}
\begin{align}
K_{i,j,r,t}=\sum_{p=\underline{p}}^{\overline{p}}\sum_{q=\underline{q}}^{\overline{q}}\sum_{s=0}^{c_2-1}K^{(1)}_{i-p,j-q,r,s}K^{(2)}_{p,q,s,t}    
\end{align}
\endgroup
\noindent where $K_{i,j,r,t}$ is the merged kernel of size $d\times d$, and $\underline{p}=\max(0,i-d_1+1)$, $\overline{p}=\min(d_2-1,i)$, $\underline{q}=\max(0,j-d_1+1)$, and $\overline{q}=\min(d_2-1,j)$. Note that when the stride of layers $s_1$ and $s_2$ is larger than 1, the kernels can still be merged into one with a stride of $s_1\times s_2$ and kernel size of $((d_2-1)\times s_1+d_1)\times((d_2-1)\times s_1+d_1)$.

\textbf{Merging inverted residual blocks}. An important insight from the above analysis is that when consecutive convolution layers are merged into one convolution layer, the number of both the input and output channels for the resulting convolution layer is only determined by the number of the input channels in the first convolution layer and the number of output channels in the last convolution layer, respectively, \textbf{regardless of the intermediate layer structures}. As a result, \textbf{the design rule of inverted residual blocks~\cite{mobilenetv2}, i.e., three convolution layers with their number of channels first expanded and then decreased, is naturally favorable to our DepthShrinker's derived networks consisting of merged convolutions with only the decreased number of input and output channels}. We believe this also sheds light on future hardware-efficient DNN designs.

\subsection{DepthShrinker$^+$: Expand-then-Shrink}
\label{sec:method_ds+}
The vanilla design of our DepthShrinker described above leverages the insight that \textit{unimportant activation functions can be properly removed after training without hurting the inference accuracy}. Excitingly, this insight can also be leveraged to improve DNN training. Specifically, we propose to train a given DNN via an Expand-then-Shrink strategy, and term it as DepthShrinker$^+$. In a DepthShrinker$^+$ training, 
we first \underline{(1)} expand one or some of the convolution layers to become inverted residual blocks, which benefits the training optimization thanks to the increased overparameterization in the expanded model, \underline{(2)} train the expanded DNN, and then \underline{(3)} apply DepthShrinker to merge the aforementioned newly introduced blocks to recover the original network structure. As such, this training scheme can be viewed as augmenting the original DNN with enhanced complexity during training to favor its training optimization and thus achievable accuracy, while the inference efficiency remains to be the same. 

%% file: Sections/5-Experiment.tex
\section{Experiment Results}
\label{sec:exp}

\subsection{Experiment Setup}
\label{sec:setup}

\textbf{Networks and datasets.} We apply DepthShrinker to both the MobileNetV2~\cite{MobilenetV1} and  EfficientNet-Lite~\cite{efficientnetlite} (i.e., a hardware-efficient variant of EfficientNet~\cite{tan2019efficientnet}) families, on top of the ImageNet dataset~\cite{russakovsky2015imagenet}.

\textbf{Search settings.} We adopt the same training hyper-parameters as the fine-tuning stage (see below), and find that the important activation functions can be quickly identified and the search becomes stable within 20 epochs.

\textbf{Fine-tuning settings.} By default, we fine-tune for 180 epochs with an SGD optimizer and a cosine learning rate, equipping with label smoothing~\cite{muller2019does} and RandAugment~\cite{cubuk2020randaugment} following~\cite{wang2021alphanet}. \textbf{Unless explicitly specified, we do not enable self-distillation in experiments of the reported results.}

\textbf{Devices and measurement settings.} We consider three commonly used computing platforms, including NVIDIA Tesla V100 GPUs~\cite{v100}, RTX 2080Ti GPUs~\cite{rtx2080ti}, and Jetson TX2 Edge GPUs~\cite{edgegpu}, and adopt the same measurement setting as in Sec.~\ref{sec:profiling}.

\begin{table}[!t]
\centering
\vspace{-0.5em}
\caption{Benchmark DepthShrinker with SOTA channel-wise pruning method AMC~\cite{he2018amc} on top of MobileNetV2@ImageNet in terms of FPS measured on three devices.}
\resizebox{0.98\linewidth}{!}
{    
\begin{tabular}{cccccc}
\toprule
\textbf{Model} & \textbf{\begin{tabular}[c]{@{}c@{}}FLOPS\\ (M)\end{tabular}} & \textbf{Acc (\%)} & \textbf{\begin{tabular}[c]{@{}c@{}}Tesla \\ V100\end{tabular}} & \textbf{\begin{tabular}[c]{@{}c@{}}RTX \\ 2080Ti\end{tabular}} & \textbf{TX2} \\ \midrule
MBV2 & 330 & 72.30 & 3088 & 2364  & 115  \\ \midrule
AMC & 220 & 70.80 & 3943  & 3159  & 152  \\
\textbf{MBV2-DS-A} & 287 & 72.43/72.48/72.50 & 5012 & 3505 & 177  \\
\textbf{MBV2-DS-B} & 272 & 71.54/72.06/72.09 & 5448  & 4074 & 199 \\
\textbf{MBV2-DS-C} & 261 & 70.90/71.21/71.56 & 6189 & 4691 & 226 \\
\textbf{MBV2-DS-D} & 253 & 69.40/70.15/70.58 & 6776 & 5257  & 258 \\ \bottomrule
\end{tabular}
}
\label{tab:exp_amc}
\vspace{-1.5em}
\end{table}

\subsection{Benchmark with SOTA Pruning Methods}
\label{sec:exp_compression}

We first benchmark DepthShrinker with SOTA pruning techniques, including both channel- and layer-wise ones.

\textbf{Benchmark with channel-wise pruning.} We benchmark with two channel-wise pruning methods, AMC~\cite{he2018amc} and MetaPruning~\cite{liu2019metapruning}, achieving SOTA performance in compressing efficient DNNs, as well as a uniform channel-wise pruning baseline in~\cite{liu2019metapruning}, on ImageNet. As shown in Tabs.~\ref{tab:exp_metapruning} and~\ref{tab:exp_amc}, we annotate our DepthShrinker's delivered model families with ``DS-X" (detailed structures are in the Appendix.~\ref{app:model_detail}), and report their accuracy under three training settings: standard training for 180 epochs, training with self-distillation in Sec.~\ref{sec:method_ft} for 180 and 360 epochs, respectively.


\underline{Results and analysis.} We can observe that \underline{(1)} under the standard training setting, DepthShrinker consistently achieves better accuracy-efficiency trade-offs over all the three baselines on all three devices. In particular, DepthShrinker achieves 1.40$\times$ throughput with a 0.18\% higher accuracy OR a 1.45\% higher accuracy with 1.14$\times$ throughput over MetaPruning-1.0$\times$, and 1.48$\times$ throughput with a 0.1\% higher accuracy over AMC on MobileNetV2 measured on a RTX 2080Ti GPU; \underline{(2)} Equipping with self-distillation and more training epochs, DepthShrinker's achievable accuracy is notably boosted by up to 2.57\%, further enlarging the accuracy gap with the channel-wise pruning baselines;
\underline{(3)} DepthShrinker shows decent scalability to different compression ratios and outperforms SOTA channel-wise pruning especially under extremely efficient cases, e.g., a 3.06\% higher accuracy with 1.53$\times$ throughput over MetaPruning-0.35$\times$ on Tesla V100; and \underline{(4)} DepthShrinker favors better data reuses and higher utilization, since the FLOPs of its delivered models are larger while their real-hardware efficiency is better compared with those of channel-wise pruning methods.

\begin{figure}[ht]
\centering
\includegraphics[width=0.93\linewidth]{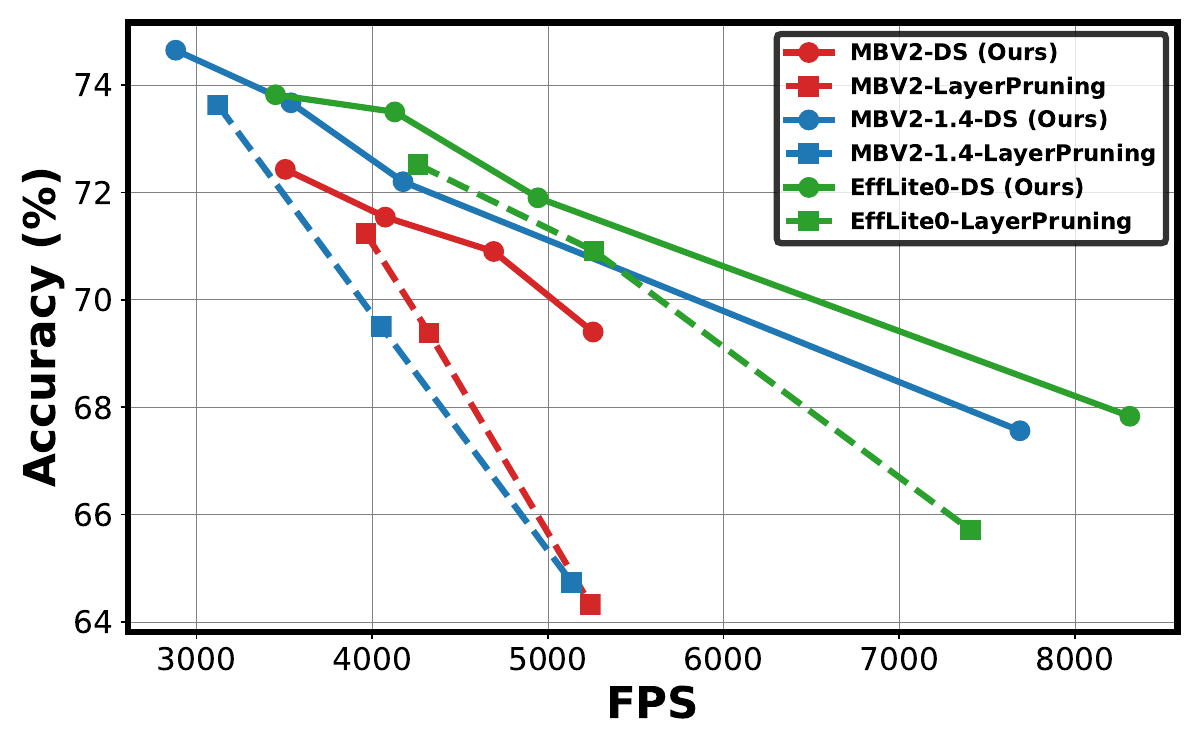}
\vspace{-1.5em}
\caption{Benchmark DepthShrinker (\textit{solid line}) with layer-wise pruning (\textit{dashed line}) on top of three models in terms of FPS measured on an RTX 2080Ti GPU. ``MBV2" and ``Efflite0" denote MobileNetV2 and EfficientNet-Lite0, respectively.}
\label{fig:layerpruning}
\vspace{-1em}
\end{figure}

\textbf{Benchmark with layer-wise pruning.} To benchmark with layer-wise pruning methods (which prune a whole block in our case), we directly remove the entire blocks identified by our DepthShrinker's differentiable search scheme for a fair comparison, based on the hypothesis that the blocks with redundant activation functions are also redundant themselves since their complexity contributes less to the final accuracy. As shown in Fig.~\ref{fig:layerpruning}, we can see that \underline{(1)} DepthShrinker still consistently achieves better accuracy-efficiency trade-offs across all three models; \underline{(2)} DepthShrinker notably outperforms layer-wise pruning under high compression ratios with more blocks pruned, since the latter suffers from a larger accuracy drop, e.g., DepthShrinker achieves a 2.80\% higher accuracy with 1.50$\times$ throughput on MobileNetV2-1.4 over the smallest model from layer pruning. This indicates merging is better than hard pruning in scalability.

\begin{figure}[ht]
\centering
\includegraphics[width=0.9\linewidth]{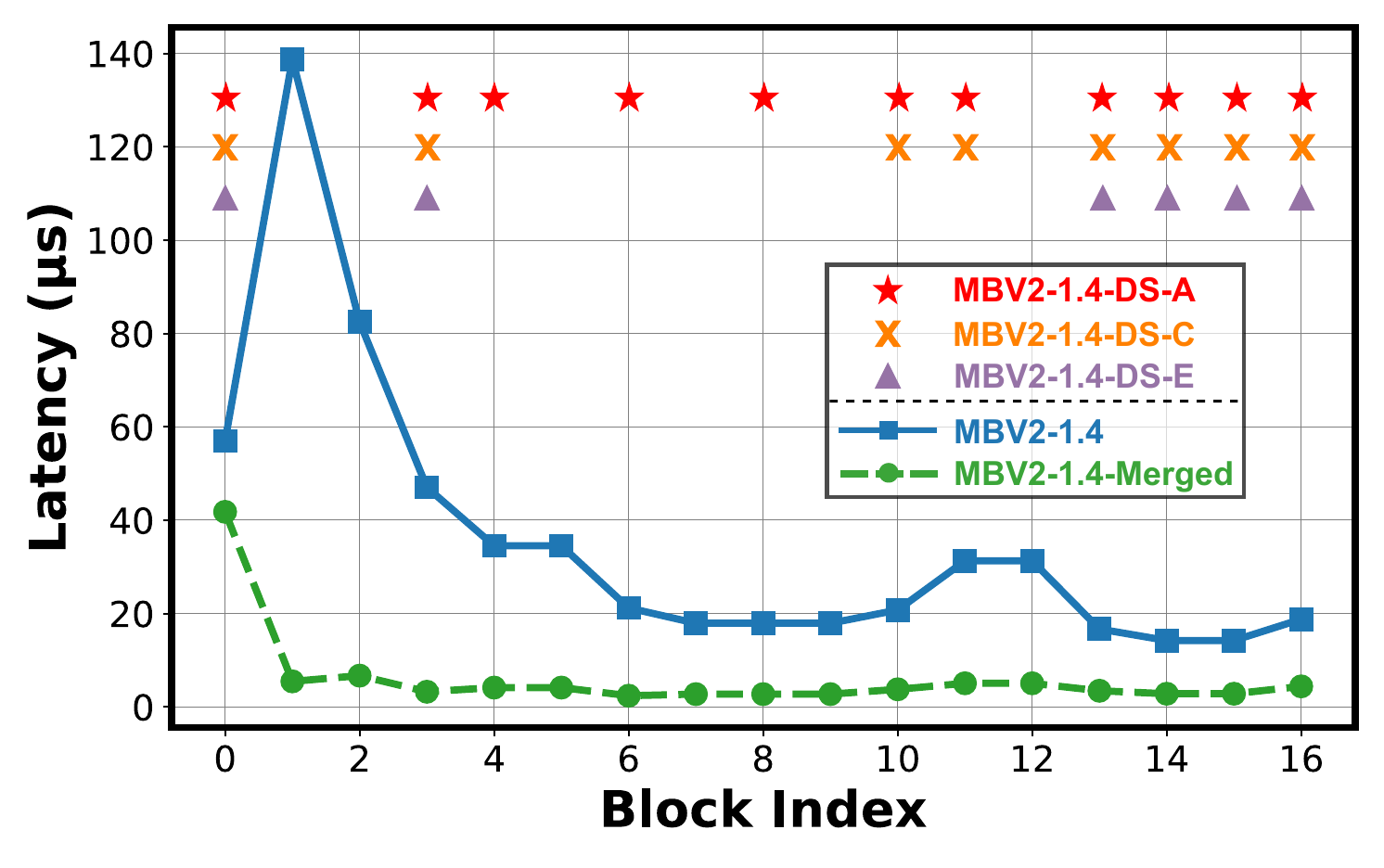}
\vspace{-1.5em}
\caption{Visualizing the block-wise latency of the blocks in MobileNetV2-1.4 (\textit{solid lines}) and their merged counterparts (\textit{dashed lines}) on an RTX 2080Ti GPU. We also annotate blocks where the activation functions are remained, using different symbols for the three model variants delivered by DepthShrinker.}
\label{fig:profile_exp}
\vspace{-0.5em}
\end{figure}

\textbf{Visualization.} We visualize the remained activation functions of DepthShrinker's delivered model variants, as well as the block-wise latency breakdown before and after merging each block on top of MobileNetV2-1.4 in Fig.~\ref{fig:profile_exp}. We can see that \underline{(1)} shrinking the building blocks to dense convolutions can notably reduce the latency by up to 96.1\%, and \underline{(2)} DepthShrinker can successfully identify bottleneck layers in terms of latency, thanks to the latency-aware decay (see Sec.~\ref{sec:method_search}). Note that a merged dense convolution has the same number of input/output channels as the original block, which is different from the setting in Sec.~\ref{sec:profiling} where the input/output channels are scaled to keep the same FLOPs.

\begin{figure}[h]
\centering
\vspace{-0.5em}
\includegraphics[width=0.95\linewidth]{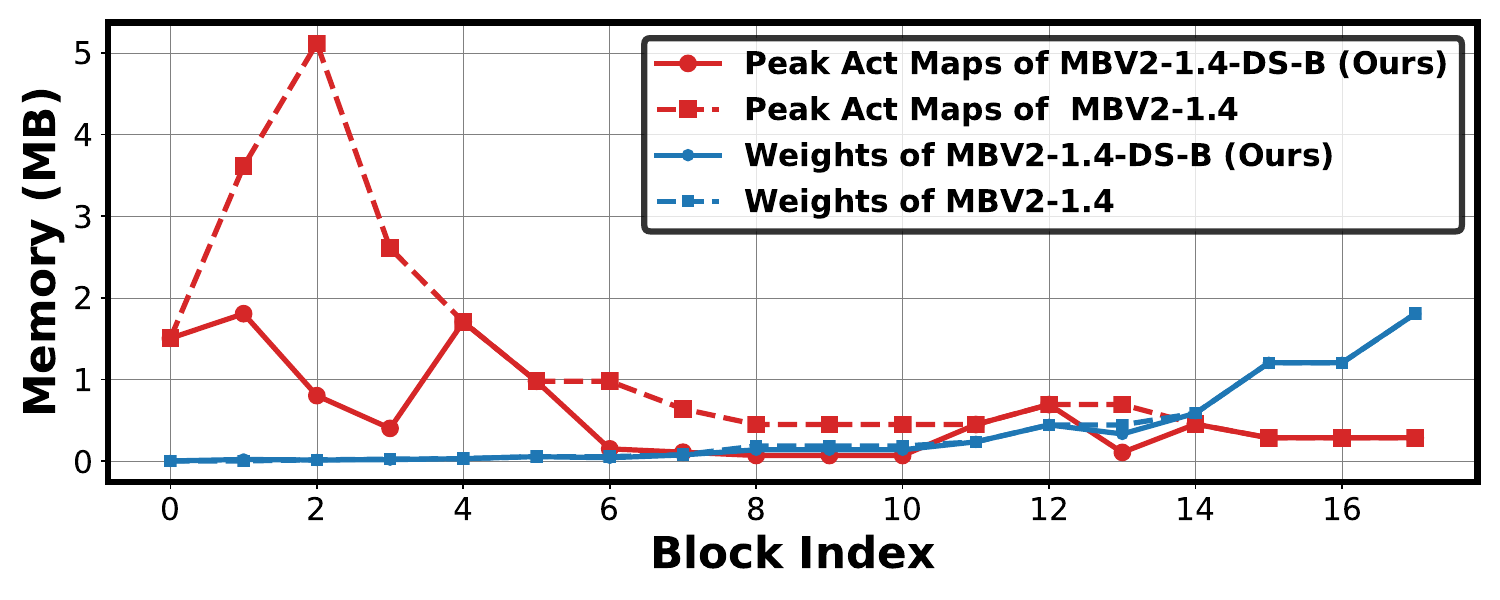}
\vspace{-1.5em}
\caption{Visualizing the memory footprint, including both that of weights and peak activation maps, of MobileNetV2-1.4 before and after applying DepthShrinker.}
\label{fig:footprint}
\vspace{-0.5em}
\end{figure}

We also visualize the block-wise memory footprint, including both that of weights and peak activation maps, i.e., the maximal sum of the input/output/residual activation maps when executing each convolution in a block, before and after applying DepthShrinker in Fig.~\ref{fig:footprint} (assuming 16-bit precision). We can see that DepthShrinker effectively reduces the peak activation usage which mostly dominates the memory footprint, as it removes both the channel expansion and residual connections, leading to reduced data movement cost and thus boosted real-hardware efficiency.

\textbf{Remark.} Our DepthShrinker opens up a new compression paradigm which provides a cost-effective perspective for compressing efficient DNN structures, wining the advantages of both channel- and layer-/block-wise pruning, i.e., achieving the high accuracy of the former together with the decent hardware efficiency of the latter.

\begin{figure}[t]
\centering
\includegraphics[width=0.9\linewidth]{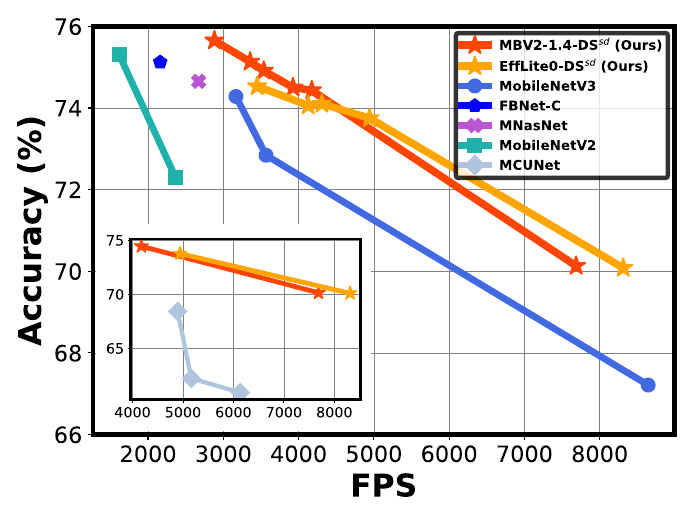}
\vspace{-1.8em}
\caption{Benchmark DepthShrinker's delivered models with SOTA efficient DNNs in terms of FPS measured on an RTX 2080Ti GPU. ``$DS^{sd}$" denotes that self-distillation is enabled. The zoom-in figure shows the comparison with MCUNet~\cite{lin2020mcunet}.}
\label{fig:sota_dnn}
\vspace{-2em}
\end{figure}

\vspace{-0.5em}
\subsection{Benchmark with SOTA Efficient DNNs}
\label{sec:exp_family}
We apply our DepthShrinker to MobileNetV2 with a channel scale of 1.4, and EfficientNet-Lite0 with self-distillation enabled to acquire a set of new model families made up of inverted residual blocks and dense convolution layers, which are compared with SOTA efficient DNN families~\cite{howard2019searching, lin2020mcunet, wu2019fbnet, lin2020mcunet, tan2019mnasnet}.

\textbf{Results and analysis.}
We show the accuracy and FPS trade-off of different models in Fig.~\ref{fig:sota_dnn}. We can observe that \underline{(1)} DepthShrinker's generated models push forward the frontier of the accuracy-efficiency trade-off over SOTA efficient DNNs, including the NAS-based ones, e.g., a 1.59\% higher accuracy with 1.17$\times$ throughput over MobileNetV3~\cite{howard2019searching}; \underline{(2)} DepthShrinker scales better to high compression ratio scenarios, e.g., a 2.87\% higher accuracy under comparable throughput (0.96$\times$) compared with the smallest model in the MobileNetV3 family.
This set of experiments indicates that shrinking manually designed models via DepthShrinker can match or even outperform advanced NAS-based models in terms of real-hardware efficiency.
Note that the key idea of DepthShrinker can be combined with NAS methods to deliver more real-hardware efficient model families, which we leave as a future work.

\subsection{Ablation study of DepthShrinker}
\label{sec:exp_ablation}
\textbf{Contributions of pretraining and adding free activation functions.} As shown in Tab.~\ref{tab:exp_ablation}, given the identified redundant activation functions, the target network is (1) trained from scratch, (2) fine-tuned from pretrained models, or (3) fine-tuned with one free activation function being added after each merged convolution (see Sec.~\ref{sec:method_ft}). We observe that \underline{(1)} pretraining improves the accuracy by 0.42\%$\sim$1.27\%, which echoes the role of activation functions in DNN training in Sec.~\ref{sec:inspiration}; and \underline{(2)} Adding the free activation functions can further boost the accuracy by up to 0.81\%.

\begin{table}[t]
\centering
\caption{Evaluating the effectiveness of starting from pretrained models and adding free activation functions.}
\resizebox{0.98\linewidth}{!}
{    
\begin{tabular}{cccc}
\toprule
\textbf{Model} & \textbf{From Scratch} & \textbf{+ Pretrain} & \textbf{+ Free Act Func} \\ \midrule
MBV2-1.4-DS-A & 73.82 & 74.40 (+0.58) & \textbf{74.65 (+0.83)} \\
MBV2-1.4-DS-C & 72.45 & 72.87 (+0.42) & \textbf{73.67 (+1.22)} \\
MBV2-1.4-DS-E & 70.12 & 71.39 (+1.27) & \textbf{72.20 (+2.08)} \\ \midrule \midrule
MBV2-DS-A & 71.51 & 72.39 (+0.88) & \textbf{72.43 (+0.92)} \\
MBV2-DS-B & 70.50 & 71.29 (+0.79) & \textbf{71.54 (+1.04)} \\
MBV2-DS-C & 69.92 & 70.51 (+0.59) & \textbf{70.90 (+0.98)} \\ \bottomrule
\end{tabular}
}
\label{tab:exp_ablation}
\vspace{-1.5em}
\end{table}

\textbf{Comparison with random search.} We benchmark the proposed differentiable search method with a random search baseline which randomly keeps the same amount of activation functions as our searched ones. Compared with the random search counterparts (averaged over five runs), 
MBV2-1.4-DS-E and MBV2-DS-D in Tabs.~\ref{tab:exp_metapruning} and~\ref{tab:exp_amc} achieve a 4.30\%/5.46\% higher accuracy with 1.24$\times$/1.19$\times$ throughput on the RTX 2080Ti GPU, respectively. This implies random search can hardly hit decent strategies and leads to both inferior accuracy and throughput. 
More ablation studies with the EfficientNet-Lite family are in the Appendix.~\ref{app:efflite}.

\textbf{Measurement on CPU devices.}
In addition to GPUs and Edge GPUs considered by aforementioned experiments,  we also measure the latency of  DepthShrinker's delivered models on two CPU devices, including the Google Pixel 3 mobile phone and Raspberry Pi 4 (Raspi 4) with a batch size of 1, where all Pytorch models are converted to ONNX and then compiled to the TFLite format, following~\cite{li2021hw}. As shown in Tab.~\ref{tab:cpu_latency}, we can see DepthShrinker still notably reduces the latency under
comparable accuracy (according to Tab.~\ref{tab:exp_metapruning} and~\ref{tab:exp_amc}), thanks to the reduced data movements with more data reuses of dense convolutions, indicating the general applicability of DepthShrinker across various commercial devices.

\begin{table}[h]
\vspace{-1.3em}
\centering
\caption{Measure the latency of DepthShrinker's delivered models on two CPU devices, i.e., Google Pixel 3 and Raspi 4.}
\resizebox{0.98\linewidth}{!}
{    
\begin{tabular}{ccc|ccc}
\toprule
\textbf{Model}  & \textbf{Pixel 3 (s)} & \textbf{Raspi 4 (s)} & \textbf{Model}  & \textbf{Pixel 3 (s)} & \textbf{Raspi 4 (s)} \\ \midrule
MBV2 & 0.073 & 0.200 & MBV2-1.4 & 0.127 & 0.299 \\ \midrule
\textbf{MBV2-DS-A} & 0.065 ($\downarrow$10.9\%) & 0.147 ($\downarrow$26.5\%) & \textbf{MbV2-1.4-DS-A} & 0.089 ($\downarrow$29.9\%) & 0.204 ($\downarrow$31.8\%) \\
\textbf{MBV2-DS-B} & 0.049 ($\downarrow$32.9\%) & 0.133 ($\downarrow$33.5\%) & \textbf{MbV2-1.4-DS-B}  & 0.105 ($\downarrow$17.3\%) & 0.196 ($\downarrow$34.4\%) \\
\textbf{MBV2-DS-C} & 0.047 ($\downarrow$35.6\%) & 0.124 ($\downarrow$38.0\%) & \textbf{MbV2-1.4-DS-C} & 0.083 ($\downarrow$34.6\%) & 0.185 ($\downarrow$38.1\%) \\
\textbf{MBV2-DS-D} & 0.045 ($\downarrow$38.4\%) & 0.116 ($\downarrow$42.0\%) & \textbf{MbV2-1.4-DS-E}  & 0.079 ($\downarrow$37.8\%) & 0.170 ($\downarrow$43.1\%) \\ \bottomrule
\end{tabular}
}
\label{tab:cpu_latency}
\vspace{-1.3em}
\end{table}

\subsection{Evaluate DepthShrinker$^+$}
\label{sec:exp_ds+}
We evaluate the proposed DepthShrinker$^+$, i.e., the Expand-Then-Shrink training strategy in Sec.~\ref{sec:method_ds+}, on top of VGG11/VGG13~\cite{simonyan2014very}/MCUNet~\cite{lin2020mcunet}/MobileNetV2~\cite{mobilenetv2} on ImageNet via replacing their intermediate blocks with inverted residual blocks, which are then merged using our DepthShrinker principle. More details about how to expand each network are in the Appendix.~\ref{app:ds+}.

\begin{table}[!h]
\centering
\vspace{-1.5em}
\caption{Evaluating DepthShrinker$^+$ on five models on ImageNet. ``rXX" denotes the input resolution, following~\cite{cai2021network}.}
\resizebox{0.98\linewidth}{!}
{    
\begin{tabular}{cccccc}
\toprule
\textbf{Model} & \textbf{VGG11} & \textbf{VGG13} & \textbf{\begin{tabular}[c]{@{}c@{}}MCUNet\\ (r176)\end{tabular}} & \textbf{\begin{tabular}[c]{@{}c@{}}MBV2-0.5\\ (r160)\end{tabular}} & \textbf{\begin{tabular}[c]{@{}c@{}}MBV2\\ (r160)\end{tabular}} \\ \midrule
Baseline (\%) & 71.51 & 71.64 & 61.50 & 61.40 & 69.60 \\
DepthShrinker$^+$ (\%) & \textbf{72.95} & \textbf{73.26} & \textbf{62.77} & \textbf{62.72} & \textbf{70.86} \\ \bottomrule
\end{tabular}
}
\label{tab:exp_ds+}
\vspace{-0.5em}
\end{table}

\textbf{Results.} As shown in Tab.~\ref{tab:exp_ds+}, DepthShrinker$^+$ consistently boosts the accuracy by 1.26\%$\sim$1.62\% over standard training across all the five models. This indicates the potential of DepthShrinker$^+$ in aiding tiny network training.

%% file: Sections/6-Conclusion.tex
 \vspace{-0.9em}
\section{Conclusion}
\label{sec:conclusion}
  \vspace{-0.5em}

To tackle the limitations of existing efficient DNNs in fulfilling their promise in boosting real-hardware efficiency due to their low hardware utilization, we open up a new compression paradigm and propose DepthShrinker to develop hardware efficient compact DNNs via merging irregular blocks into dense operations with much improved real-hardware efficiency. Extensive experiments validate our DepthShrinker wins both the high accuracy of channel-wise pruning and the decent efficiency of layer-wise pruning, opening up a cost-effective dimension for DNN compression.

\vspace{-0.5em}
\section*{Acknowledgements}

The work performed by Yonggan Fu, Jiayi Yuan, Cheng Wan, and Yingyan Lin is supported by the National Science Foundation (NSF) through the SCH program (Award number: 1838873), the NeTS program (Award number: 1801865), and the MLWiNS program (Award number: 2003137).

%% file: Sections/7-Appendix.tex
\newpage
\appendix

\section{Evaluate DepthShrinker on EfficientNet-Lite Families}
\label{app:efflite}

\textbf{Setup.} We apply DepthShrinker on top of EfficientNet-Lite3~\cite{efficientnetlite} on ImageNet, equipped with the self-distillation mechanism mentioned in Sec.~\ref{sec:method_ft}, to generate a new model family annotated as ``EffLite3-DS-X" in Tab.~\ref{tab:exp_efflite}. For a fair comparison, we adopt the same training schedule as introduced in Sec.~\ref{sec:setup} to train the EfficientNet-Lite baselines from scratch. 

\textbf{Results and analysis.} From Tab.~\ref{tab:exp_efflite}, we can observe that DepthShrinker's delivered models again push forward the achievable accuracy-efficiency trade-off. In particular, EffLite3-DS-B achieves a 1.32\% higher accuracy with 1.21$\times$ throughput on Tesla V100 over EfficientNet-Lite2 and EffLite3-DS-A achieves a 1.38\% higher accuracy with comparable throughput (e.g., 0.95$\times$ on Tesla V100). As mentioned in the main text, our DepthShrinker can be combined with NAS methods to deliver new model families featuring much improved real-hardware efficiency and we leave this as our future work.

\begin{table}[!ht]
  \vspace{-1em}
\centering
\caption{Evaluating DepthShrinker on top of EfficientNet-Lite3 on ImageNet. ``Efflite" denotes EfficientNet-Lite.}
\resizebox{0.98\linewidth}{!}
{    
\begin{tabular}{cccccc}
\toprule
\multirow{2}{*}{\textbf{Model}} & \multirow{2}{*}{\textbf{Acc (\%)}} & \multirow{2}{*}{\textbf{MFLOPs}} & \multicolumn{3}{c}{\textbf{Throughput}} \\
 &  &  & \textbf{Tesla V100} & \textbf{RTX 2080Ti} & \textbf{TX2} \\ \midrule
EffLite1 & 75.41 & 651 & 1773 & 1403 & 64 \\
EffLite2 & 76.14 & 924 & 1301 & 1081 & 44 \\ \midrule
\textbf{EffLite3-DS-A} & 76.79 & 991 & 1676 & 1337 & 53 \\
\textbf{EffLite3-DS-B} & 77.46 & 952 & 1573 & 1264 & 50 \\
\textbf{EffLite3-DS-C} & 77.63 & 918 & 1431 & 1152 & 46 \\
\textbf{EffLite3-DS-D} & 77.84 & 905 & 1250 & 1030 & 41 \\ \bottomrule
\end{tabular}
}
\label{tab:exp_efflite}
\vspace{-1em}
\end{table}

\section{More Benchmark with Layer-wise Pruning}

\begin{figure}[ht]
\centering
\includegraphics[width=0.9\linewidth]{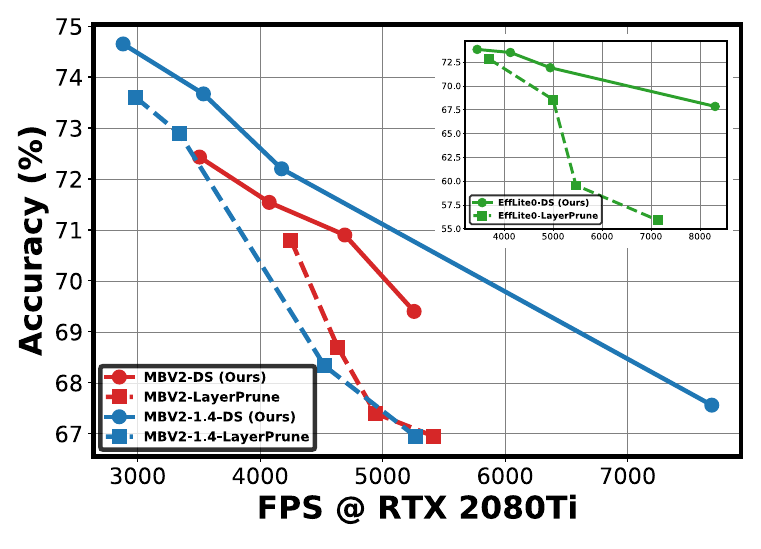}
\vspace{-1.5em}
\caption{Benchmark DepthShrinker (\textit{solid line}) with LayerPrune~\cite{elkerdawy2020filter} (\textit{dashed line}) on top of three models in terms of FPS measured on an RTX 2080Ti GPU.}
\label{fig:layerprune}
\vspace{-1em}
\end{figure}

We also benchmark with LayerPrune~\cite{elkerdawy2020filter} for compressing three efficient models based on their provided implementation, as a complement of Sec.5.2 in the main text. As shown in Fig.~\ref{fig:layerprune}, DepthShrinker still consistently outperforms LayerPrune, especially under large compression ratios.

\section{More Real-device Profiling Results}
\label{app:profile}

We  benchmark the efficiency of SOTA DNN families under different depth/width trade-offs under the same FLOPs for better understanding the causes of the improved hardware utilization in Sec.~\ref{sec:profiling}.

\textbf{Setup.}  
To construct models featuring different depth/width trade-offs with the same FLOPs, we uniformly scale the channel number of the networks within the same model family and benchmark their throughput on different devices.

\begin{table*}[ht]
  \vspace{-0.3em}
\centering
\caption{Measured throughput of the ResNet family and the MobileNetV2 family with scaled channel numbers to maintain the same FLOPs. All the reported numbers are FPS. The number of blocks in each stage of MobileNetV2 is annotated in the Depth column.}
\resizebox{0.85\linewidth}{!}
{
\begin{tabular}{c|cc|ccc}
\toprule
\multirow{2}{*}{\textbf{Model}} & \multirow{2}{*}{\textbf{Depth}} & \multirow{2}{*}{\textbf{Width Scale}} & \multicolumn{3}{c}{\textbf{Throughput (FPS)}} \\
 &  &  & \textbf{RTX 2080Ti} & \textbf{TX2 (bs=32)} & \textbf{TX2 (bs=1)} \\ \midrule
ResNet-18 & 18 & 1.535 & 1476 & 85 & 28 \\
ResNet-34 & 34 & 1.07 & 1388 & 61 & 27 \\
ResNet-50 & 50 & 1.00 & 874 & 46 & 20 \\
ResNet-101 & 101 & 0.73 & 792 & 36 & 15 \\
ResNet-152 & 152 & 0.60 & 674 & 31 & 11 \\ \midrule \midrule
\multirow{5}{*}{MobileNetV2} & {[}1,1,1,1,1,1,1{]} & 1.45 & 2499 & 112 & 56 \\
 & {[}1,2,2,2,1,1,1{]} & 1.25 & 2167 & 98 & 52 \\
 & {[}1,2,3,3,2,2,1{]} & 1.11 & 2114 & 103 & 39 \\
 & {[}1,2,3,4,3,3,1{]} & 1.00 & 2149 & 103 & 33 \\
 & {[}1,3,4,6,5,5,1{]} & 0.85 & 1916 & 96 & 23 \\ \bottomrule
\end{tabular}
}
\label{tab:profile_depth_width}
\vspace{-1em}
\end{table*}

\textbf{Results and analysis.}
As shown in Tab.~\ref{tab:profile_depth_width}, we can observe that \underline{(1)} shallower networks consistently win better throughput compared with the deeper counterparts under the same FLOPs across different model families and devices, e.g., channel-scaled ResNet-50 achieves 1.30$\times$ over channel-scaled ResNet-152. This indicates the preference for shallow-wide networks of the mapping strategies of existing commercial devices; \underline{(2)} shallower networks  reduce the latency by up to 58.9\% on TX2 Edge GPU (i.e., the inverse of the throughput measured with a batch size of one), which is another perspective for measuring the real-time processing capability on edge devices; \underline{(3)} Based on the comparison between Tab.~\ref{tab:profile_depth_width} and Tab.~\ref{tab:profiling}, dense operations win real-hardware efficiency thanks to both of the two aforementioned aspects and reducing operation-wise irregularity may contribute more, especially on more powerful devices with a higher degree of parallelism.

\section{Design Details of DepthShrinker$^+$}
\label{app:ds+}

In Sec.~\ref{sec:exp_ds+},  we evaluate the proposed DepthShrinker$^+$, on top of VGG11/VGG13~\cite{simonyan2014very}/MCUNet~\cite{lin2020mcunet}/MobileNetV2~\cite{mobilenetv2} on ImageNet. In particular, for the last two models, we follow the definition in~\cite{cai2021network}. Without bells and whistles, we design empirical rules to determine which layer to be expanded for all the networks to demonstrate the general effectiveness of our DepthShrinker technique as a training technique for boosting accuracy. 

\textbf{Expand VGG.} For all VGG networks, we expand all the 3$\times$3 convolution layers to a standard inverted residual block~\cite{mobilenetv2} with an expansion ratio of 6, except the first two convolution layers and the last convolution layer.

\textbf{Expand MCUNet/MobileNetV2.} For all networks made up of inverted residual blocks, we apply DepthShrinker$^+$ to expand one block in every two consecutive blocks. In addition, to expand one specific inverted residual block, we only expand the first pointwise convolution to a new inverted residual block~\cite{mobilenetv2} with an expansion ratio of 6 and a depthwise kernel size of 1 to ensure the original model structure can be recovered.

Integrating with more advanced expansion strategies, our DepthShrinker can potentially achieve more notable improvements, which will be our future work.

\begin{table}[ht]
  \vspace{-1em}
\centering
\caption{Visualizing the remained activation functions in DepthShrinker's generated model families.}
\resizebox{0.98\linewidth}{!}
{    
\begin{tabular}{cc}
\toprule
\textbf{Model} & \textbf{Remained Activation Functions} \\ \midrule
MBV2-1.4-DS-A & {[}1 0 0 1 1 0 1 0 1 0 1 1 0 1 1 1 1{]} \\
MBV2-1.4-DS-B & {[}0 0 0 1 1 0 0 0 0 0 1 1 0 1 1 1 1{]} \\
MBV2-1.4-DS-C & {[}1 0 0 1 0 0 0 0 0 0 1 1 0 1 1 1 1{]} \\
MBV2-1.4-DS-D & {[}0 0 0 1 1 0 0 0 0 0 0 0 0 1 1 1 1{]} \\
MBV2-1.4-DS-E & {[}1 0 0 1 0 0 0 0 0 0 0 0 0 1 1 1 1{]} \\
MBV2-1.4-DS-F & {[}0 0 0 0 0 0 0 0 0 0 0 0 0 0 0 0 0{]} \\ \midrule \midrule
MBV2-DS-A & {[}0 0 1 0 1 1 1 0 0 1 1 1 1 1 1 1 1{]} \\
MBV2-DS-B & {[}1 0 0 0 1 1 1 0 0 1 1 1 1 1 1 1 1{]} \\
MBV2-DS-C & {[}1 0 0 1 1 0 1 0 0 0 1 0 0 1 1 1 1{]} \\
MBV2-DS-D & {[}1 0 0 1 0 0 0 0 0 0 1 1 0 1 0 1 1{]} \\ \midrule \midrule
Eff-Lite0-A & {[}0 0 1 1 0 1 0 0 1 0 0 1 1 1 1 1{]} \\
Eff-Lite0-B & {[}0 0 0 1 1 0 0 0 1 0 0 1 1 1 1 1{]} \\
Eff-Lite0-C & {[}0 0 0 1 0 0 0 0 0 0 0 1 1 1 1 1{]} \\
Eff-Lite0-D & {[}0 0 0 0 0 0 0 0 0 0 0 0 0 0 0 0{]} \\ \midrule \midrule
EffLite3-DS-A & {[}0 0 0 1 1 0 0 0 0 0 0 0 0 0 0 1 0 0 1 1 1 1 1 1{]} \\
EffLite3-DS-B & {[}0 0 0 1 1 0 0 0 0 0 0 0 1 1 0 0 0 1 1 1 1 1 1 1{]} \\
EffLite3-DS-C & {[}0 0 0 1 1 0 0 0 0 0 1 0 1 1 1 0 0 1 1 1 1 1 1 1{]} \\
EffLite3-DS-D & {[}1 0 0 1 1 0 1 1 0 0 0 0 1 0 0 1 1 1 1 1 1 1 1 1{]} \\ \bottomrule
\end{tabular}
}
\label{tab:model_detail}
\vspace{-1em}
\end{table}

\section{Details about DepthShrinker's Delivered Model Families}
\label{app:model_detail}

We apply DepthShrinker on top of the given efficient DNNs to generate new model families via varying the number of remained activation functions $k$ in Eq.~\ref{eq:search_objective} and the decay strength on $m$ discussed in Sec.~\ref{sec:method_search}, which constrains the overall efficiency of the delivered network. For all the reported models in the main text, we provide their remained activation functions identified by our DepthShrinker in Tab.~\ref{tab:model_detail}, where each element in the list indicates whether the activation functions of the corresponding block are kept, i.e., ``1" denotes the activation functions in the block are remained.